\documentclass{article}
\usepackage[total={6.5in, 9in}]{geometry}
\usepackage{authblk}
\usepackage{multicol}
\usepackage{bm} 
\usepackage{multicol}
\usepackage{amsmath}
\usepackage{hyperref}
\usepackage[inline]{enumitem}
\usepackage[english]{babel}
\usepackage{todonotes}
\usepackage{xcolor}
\usepackage{color,soul}
\usepackage{booktabs}
\usepackage{dirtytalk}

\definecolor{blue}{HTML}{1F77B4}
\definecolor{purple}{HTML}{7a00a3}
\definecolor{green}{HTML}{00a35f}
\hypersetup{colorlinks, citecolor=blue, urlcolor=blue, linkcolor=blue}
\definecolor{orcidcolor}{HTML}{A6CE39}
\usepackage{csquotes}
\usepackage[style=authoryear, backend=biber, bibencoding=utf8,
    defernumbers=true, maxbibnames=99, maxcitenames=3, language=english,
]{biblatex}
\DeclareCiteCommand{\cite}
  {\usebibmacro{prenote}}
  {\usebibmacro{citeindex}%
   \printtext[bibhyperref]{\usebibmacro{cite}}}
  {\multicitedelim}
  {\usebibmacro{postnote}}

\DeclareCiteCommand*{\cite}
  {\usebibmacro{prenote}}
  {\usebibmacro{citeindex}%
   \printtext[bibhyperref]{\usebibmacro{citeyear}}}
  {\multicitedelim}
  {\usebibmacro{postnote}}

\DeclareCiteCommand{\parencite}[\mkbibparens]
  {\usebibmacro{prenote}}
  {\usebibmacro{citeindex}%
    \printtext[bibhyperref]{\usebibmacro{cite}}}
  {\multicitedelim}
  {\usebibmacro{postnote}}

\DeclareCiteCommand*{\parencite}[\mkbibparens]
  {\usebibmacro{prenote}}
  {\usebibmacro{citeindex}%
    \printtext[bibhyperref]{\usebibmacro{citeyear}}}
  {\multicitedelim}
  {\usebibmacro{postnote}}

\newcommand{\orcidhref}[3][orcidcolor]{\href{#2}{\color{#1}{#3}}}
\newcommand{\pporcid}[1]{
        \hspace{5mm}\textcolor{orcidcolor}{} \orcidhref{https://orcid.org/#1}{ORCID: #1}
}
\def\email#1{{\tt#1}}
\makeatletter
\let\blx@rerun@biber\relax
\makeatother
\begin{document}


\title{Unsigned Play by Milan Kundera? \\ An Authorship Attribution Study}
\author[1]{Lenka Jungmannová}
\author[2]{Petr Plecháč}

\affil[1]{Institute of Czech Literature, Czech Academy of Sciences, Prague, Czech Republic
    \protect\\[1mm]
    \makebox[4.5cm][l]{\email{jungmannova@ucl.cas.cz}}  
    \makebox[4.5cm][r]{\pporcid{0000-0003-1704-2023}}
} 

\affil[2]{Institute of Czech Literature, Czech Academy of Sciences, Prague, Czech Republic
    \protect\\[1mm]
    \makebox[4.5cm][l]{\email{plechac@ucl.cas.cz}}  
    \makebox[4.5cm][r]{\pporcid{0000-0002-1003-4541}}
} 
\date{}
\maketitle          


\begin{abstract}
\noindent In addition to being a widely recognised novelist, Milan Kundera has also authored three pieces for theatre: \textit{The Owners of the Keys} (\textit{Majitelé klíčů}, 1961), \textit{The Blunder} (\textit{Ptákovina}, 1967), and \textit{Jacques and his Master} (\textit{Jakub a jeho pán}, 1971). In recent years, however, the hypothesis has been raised that Kundera is the true author of a fourth play: \textit{Juro Jánošík}, first performed in a 1974 production under the name of Karel Steigerwald, who was Kundera’s student at the time. In this study, we make use of supervised machine learning to settle the question of authorship attribution in the case of \textit{Juro Jánošík}, with results strongly supporting the hypothesis of Kundera’s authorship.
\end{abstract}
\vspace{1cm}

\begin{multicols}{2}

\section{Introduction}

Of Milan Kundera’s plays, there is only one that is generally known abroad: an adaptation of Diderot’s novel \textit{Jacques and His Master} completed by Kundera in 1971, and later included as the only play in a collection of his works \parencite[625--698]{kundera2011}. In the territory of the former Czechoslovakia, however, Kundera is recognised as a full-fledged playwright, and author of three plays. In this paper we present the evidence that Kundera is most likely the author of a fourth play, one that has typically been attributed to a different author (under whose name the play was first produced).

\section{History and Related Works}

When the director Otomar Krejča was hired as new head of drama at the Prague National Theatre in 1956, he and dramaturge Karl Kraus created a workshop for authors which they eventually invited Kundera to join in 1959. It is in this context that Kundera in 1961 completed his first play \textit{The Owners of the Keys} (\textit{Majitelé klíčů}), though it would premiere elsewhere, after it was initially decided that a production at the National Theatre would not be permitted (the decision was eventually reversed and the play permitted at the National Theatre following productions at three other theatres).

In line with totalitarian ideology, the play mocks the petty bourgeois for refusing to accept socialist doctrine. Set during the Nazi occupation of Bohemia and Moravia, it portrays an elderly couple as they engage in petty arguments, praise the ownership class, and (most notably) blindly acquiesce to the occupation, in this way helping to keep Nazism in power. While their daughter agrees with their professed capitalist views, their son-in-law Jiří, together with Věra (a resistance fighter who comes under pretence of a visit, but is secretly fleeing the Gestapo), present the idealised image of communists biding their time until the war ends to seize political power---something that actually came to pass in 1948. The moral problem of the play is introduced through Jiří’s murder of a janitor who is spying on Věra for the Nazis. Following the incident, he and Věra decide to go underground, thus leaving the family to almost certain death. Alongside the main narrative, the audience is presented with ghostly, psychoanalytically conceived montage scenes that serve to underscore the plot.

Despite the play’s predominantly tendentious artistic statement, it also features a mildly controversial criticism of Stalinism, and touches on the theme of seduction. These elements represent a striking departure from theatrical conventions of the time, as did also Kundera’s unique dramatic device of \say{simultaneous dialogue}, in which two different conversations are delivered side by side on stage, producing certain moments of mutual resonance. In 1990, Kundera prohibited all further productions of the play.

In 1967, Kundera completed his second play \textit{The Blunder} (\textit{Ptákovina}), originally produced under the title \textit{Two Ears, Two Weddings} (\textit{Dvě uši, dvě svatby}). By the second half of the 1960s, Czechoslovakia was undergoing a process of democratisation, a changing historical context that also affected Kundera’s work on \textit{The Blunder}---the work of a playwright no longer required to toe the line of regime ideology. Unfortunately, the period was brought abruptly to an end on 21 August 1968 with the invasion and subsequent occupation of Czechoslovakia by Warsaw Pact troops. Even then, however, three different productions of Kundera’s controversial play were successfully mounted in 1969.

A satire in the Orwellian tradition, \textit{The Blunder} sends up totalitarianism by caricaturing the mechanism it uses to blackmail the people. A school headmaster draws the Czech \say{pictogram} for the female genitalia on a blackboard: it is a mockery of the rules dictated by power, which he professes but does not himself observe. He is followed by his teacher Eva, who falls in love with him. An investigation is opened and a dogmatic official of the district authority comes to the school. An innocent student confesses to the crime, and the official demands that his ears be cut off and that Eva, as his teacher, is to be whipped. Subsequently, the headmaster is recruited by the official to test the fidelity of his fiancée. The headmaster seduces her, but reports back to the official that she did not succumb. The play does not end happily, however, and after the absurd portrayal of one of two weddings, it is revealed that the fiancée has recordings of the headmaster insulting the official. Using these recordings, she proceeds to blackmail the headmaster into having sex with her, using sexuality in this sense to dominate her partners, and indirectly the whole of society. Here too, Kundera uses the device of \say{simultaneous dialogue} – and here too, in 1990, he prohibited all further productions of the play.

Following \textit{The Blunder}, the only play to which Kundera claims authorship is his adaptation of Diderot’s novel \textit{Jacques and His Master}, commissioned by the Drama Theatre of Ústí nad Labem. By this time, he had already been removed from public life, so the director Evald Schorm was called on to sign the play in his place. (This use of pseudonyms or \say{substitute authors} was a common but clandestine aspect of literary life in totalitarian Czechoslovakia.)

The plot of \textit{Jacques and His Master} consists of little more than the aimless wandering of a servant and his master who give air to commonly held ideas, and then illustrate them with stories of their past romantic adventures: Jakub recalls how he lost his virginity, the master speaks on his unrequited love, and the innkeeper recounts the story of an acquaintance who took revenge on her noble lover by driving him into marriage with a prostitute. In this play as well, we find Kundera dividing space-time into two discrete parts as well as the device of \say{;simultaneous dialogue}, to depict either the characters in the course of telling their stories, or the dramatisation of these stories in the form of a \say{play within a play}. The stories conclude at last with the heroes confessing that through their various love affairs they have sired illegitimate sons, meanwhile coming to the understanding that the journey is more important than the destination. In addition to demonstrating their freedom, the characters refute the communist notion of linear progress with the especially subversive notion that \say{Forward – that’s anywhere. […] Wherever you happen to look is straight forward!} (\cite[697]{kundera2011}; \cite[113]{kundera1997}). 

Today, the prevailing opinion among experts is that Kundera is the author of at least one other play, namely \textit{Juro Jánošík} (1973), which was signed by his former student, the playwright Karel Steigerwald.\footnote{This hypothesis was formulated on the basis of a comparison of authorial poetics (see \cite{jungmannova2017}). Kundera’s biographer Jan Novák makes a similar claim \parencite{novak2020}.}  Even the playwright Jana Knitlová, who prepared a 1974 production of the play at the Jiří Wolker Theatre, has refused to comment on its authorship.

Juro Jánošík is a dramatic adaptation of the real-life Slovak legend of a bandit who stole from the rich and gave to the poor---and who thus also came to figure in propaganda for the Communist regime. In order for the play to pass in the socialist theatre system, the author was compelled to frame the material in pro-communist terms: the hero of the story chooses to live as a bandit in order to defend serfs from the feudalistic injustice of their masters. (Under \say{normalisation}, all literary characters not explicitly portrayed as proletarian were given a negative class label.) The plot unfolds around plans for the marriage of the Countess’s daughter to a merchant. Learning of this, the bandits decide to attend the wedding in disguise. This compels the tireless military colonel to blackmail the girlfriend of one of the bandits, through which he learns of the group’s next meeting and eventually captures Jánošík. Brave to the end, the hero is offered a place in the army, but chooses death instead.

The author sets himself a difficult task in this play. While it seems to conform to the ideology of the Communist regime, it also serves as a condemnation of the Soviet occupation. We see the latter in certain allegorical references: the Countess’s attitude towards the army stationed in her castle, for example, and in her complaint that \say{They will go on protecting us until we are completely broke} (\say{Ti nás budou tak dlouho chránit, až budeme úplně na mizině}; \cite[26]{steigerwald1974}.).  We also find self-referential allusions in the play that are characteristic of Kundera’s poetics---in Ilčík’s response to Jánošík’s question of whether his friends know any of the bandits who got away, for example: \say{They too were like that. Today they live somewhere in peace, under a foreign name, recalling their happy youth} (\say{Taky takoví byli. Dnes si žijí někde v klidu pod cizím jménem a vzpomínají na veselé mládí}; \cite[8]{steigerwald1974}). 

There are, however, other strikingly \say{Kunderian} aspects of Juro Jánošík. A comparison of dramatic forms featured in the plays named above seems particularly convincing: in all cases they take the form of dramatic montage, within which parts that differ in kind (for example, individual plot levels) are composed into a meaningful new whole. We see further similarities in the unique treatment of space-time, with scenes presented simultaneously, organised either left-right (side by side), front-back, or top-bottom. This dramatic conceit includes the device of \say{simultaneous dialogue} mentioned above, in which two different conversations are delivered side by side on stage to produce moments of mutual resonance.

\section{Materials and Methods}

To test the hypothesis of Kundera’s authorship of \textit{Juro Jánošík}, we perform a stylometric analysis. Our corpus consists of nine documents: the target text, the three plays known to be authored by Kundera, four of Steigerwald’s earliest plays (written between 1979 and 1982), and Steigerwald’s radio play of 1976 (see \autoref{tab:corpus} for details; from now on we refer to these plays by their \say{short title} in column 5). With the exception of this last, all texts have been machine-transcribed from professional editions using OCR technology. For Steigerwald’s radio play, as it was never published in print, and as the manuscript is not available, we created a machine-transcription from an audio recording in the Czech Radio Archive \parencite{steigerwald1976} using the Google Cloud Speech-To-Text API \parencite{googleSTT}.\footnote{None of our features relies directly on punctuation so we have decided not to force it into transcription of the text. This may theoretically cause some issues with lemmatisation, yet an experiment kindly performed by Milan Straka with PDT data \parencite{pdt2018} shows no significant drop in accuracy when punctuation is omitted (accuracy with original texts: 0.9928; accuracy without punctuation: 0.9910).} We have manually corrected all texts for misrecognised characters, as well as removing character names, stage directions, songs, and quotations of other works. Plain dialogues were then tokenised and lemmatised by means of \texttt{UDPipe 2} \parencite{straka2018}. To increase the dataset size, each text was split into consecutive chunks of 2000 words (samples).

\begin{table*}[ht]
\small
\center
\begin{tabular}{lllllcc}
\toprule
author & year & Czech title & English title & short title & type & \# of samples\\
\midrule
Kundera & 1962 & Majitelé klíčů & The Owners of the Keys & Owners & T & 4\\
 & 1968 & Ptákovina & The Blunder & Blunder & T & 4\\
 & 1971 & Jakub a jeho pán & Jacques and his Master & Jacques & T & 5\\
\textit{disputed} & 1974 & Juro Jánošík & Juro Jánošík & Jánošík & T & 3\\
Steigerwald & 1976 & Slabé odpolední slunce & Pale Afternoon Sun & Sun & R & 2\\
 & 1979 & Tatarská pouť & The Tatar Fair & Fair & T & 5\\
 & 1980 & Dobové tance & The Period Dances & Dances & T & 4\\
 & 1980 & Foxtrot & Foxtrot & Foxtrot & T & 5\\
 & 1982 & A tak tě prosím, kníže... & And so I ask you, prince... & Prince & T & 2\\
\toprule
\end{tabular}
\caption{Corpus details. In „type“ column „T“ stands for theatre play, „R“ stands for radio play}
    \label{tab:corpus}
\end{table*}

As sample representations we use lemmata frequencies, word frequencies, frequencies of character trigrams, and lemmata + character trigram frequencies combined. (In a rich inflection language such as Czech, these combined frequencies should theoretically capture authorial signals from two different levels: vocabulary and morphology.) Relative frequencies are transformed to \textit{z}-scores and data is trimmed at four different levels: 500, 750, 1000, and 1500 most frequent types (MFT). This results in 16 different datasets.

We employ two different methods: (1) hierarchical agglomerative clustering based on cosine distance and (2) supervised machine learning, namely \textit{Support Vector Machine} (SVM).\footnote{For the sake of compatibility between both methods used, we employ SVM with cosine similarity kernel, instead of more common linear kernel.} The former is usually referred to in stylometry as the Cosine Delta Measure \parencite{smith2011}. Both of these methods have proven reliable in a wide range of authorship attribution tasks \parencite{savoy2020}.

\section{Results}

We evaluate our SVM models by means of cross-validation in the following way: in each round we take samples from a single play as a test set and the rest of the samples as a training set. The training set thus never contains samples from a play that is undergoing classification at that moment. Since SVM is generally sensitive to imbalanced data, in each round we randomly lower the number of Steigerwald’s samples in training set to the number of Kundera’s samples. This procedure is repeated 1,000 times for all 16 datasets, i.e. each sample is classified 16,000 times.

\autoref{tab:svm-acc} shows that while overall accuracy is fair (0.94), it is not distributed evenly across particular plays and datasets. As for \textit{theatre} plays, misclassifications concern almost exclusively Kundera’s samples: accuracy estimations for Kundera’s plays range from 0.64 to 1.0 across individual datasets, with mean values between 0.85 and 0.95, while Steigerwald’s samples are always assigned to their true author, with the exception of one case. The highest error rate, however, is found with Steigerwald’s \textit{radio} play (\textit{Sun}). Here individual datasets score between 0.4 and 0.96 with a mean value of 0.76. The differences between particular features and MFT levels are rather low. In general, the combination of lemmata + trigrams yields higher accuracy than other features, but (1) its addition as compared to sole lemmata is negligible and (2) its accuracy is not consistently higher across particular plays.

\setlength\tabcolsep{2pt}
\begin{table*}[ht]
\small
\center
\begin{tabular}{llccccccccccccccccc}
\toprule
 &  & \multicolumn{4}{c}{words} & \multicolumn{4}{c}{lemmata} & \multicolumn{4}{c}{trigrams} & \multicolumn{4}{c}{lemmata + trigrams} \\
\cmidrule(lr){3-6} 
\cmidrule(lr){7-10}
\cmidrule(lr){11-14}
\cmidrule(lr){15-18}
author & play & 500 & 750 & 1000 & 1500 & 500 & 750 & 1000 & 1500 & 500 & 750 & 1000 & 1500 & 500 & 750 & 1000 & 1500 & \textit{all} \\
\midrule

Kundera & Owners  &0.97&0.88&0.77&0.74&0.84&0.87&0.82&0.71&0.76&0.84&0.82&0.89&0.92&0.94&0.94&0.97&\textit{0.85}  \\
& Blunder  &0.99&0.99&0.96&0.82&0.96&0.94&0.87&0.95&0.97&0.99&0.97&0.9&0.98&0.99&0.95&0.94&\textit{0.95}  \\
& Jacques  &0.76&0.81&0.9&0.88&0.98&1.0&0.98&0.93&0.64&0.78&0.79&0.88&0.83&0.94&0.92&0.95&\textit{0.87}  \\
Steigerwald & Sun  &0.4&0.51&0.9&0.51&0.6&0.95&0.87&0.88&0.96&0.86&0.88&0.78&0.79&0.77&0.8&0.75&\textit{0.76}  \\
& Fair  &1.0&1.0&1.0&1.0&1.0&1.0&1.0&1.0&0.8&1.0&1.0&1.0&1.0&1.0&1.0&1.0&\textit{0.99}  \\
& Dances  &1.0&1.0&1.0&1.0&1.0&1.0&1.0&1.0&1.0&1.0&1.0&1.0&1.0&1.0&1.0&1.0&\textit{1.0}  \\
& Foxtrot  &1.0&1.0&1.0&1.0&1.0&1.0&1.0&1.0&1.0&1.0&1.0&1.0&1.0&1.0&1.0&1.0&\textit{1.0}  \\
& Prince  &1.0&1.0&1.0&1.0&1.0&1.0&1.0&1.0&1.0&1.0&1.0&1.0&1.0&1.0&1.0&1.0&\textit{1.0}  \\
\multicolumn{2}{c}{\textit{all}} &\textit{0.92} &\textit{0.92} &\textit{0.94} &\textit{0.89} &\textit{0.95} &\textit{0.97} &\textit{0.95} &\textit{0.94} &\textit{0.87} &\textit{0.93} &\textit{0.93} &\textit{0.94} &\textit{0.95} &\textit{0.97} &\textit{0.96} &\textit{0.96} &\textit{0.94}  \\

\toprule
\end{tabular}
\caption{Cross-validation of SVM models (repeated 1,000 times). Accuracy estimations for particular plays and different datasets}
    \label{tab:svm-acc}
\end{table*}
\setlength\tabcolsep{6pt}

We now turn to the classification of \textit{Jánošík}. Here we use samples from all eight undisputed plays as a training set and apply the same procedure as above (lowering the number of Steigerwald’s samples in 1,000 iterations). The results are unequivocal: all 16,000 classification runs predict Kundera as an author. Given the high accuracy of our models, such a clear outcome seems very convincing. Before accepting it, however, we need to consider two possible factors.

As we have seen, \textit{Sun} is more prone to misclassifications than the other plays. We may consider three hypotheses here: 

\begin{enumerate}[label=(H\arabic*)]
\item they are due to genre discrepancies between training and test sets (theatre plays / radio play),
\item they are due to the fact that Steigerwald’s authorial style was obscured by the interference of the script editor of the recording (Vladimír Gromov), or
\item they are due to Steigerwald’s very early style being strongly influenced by Kundera.
\end{enumerate}

Should (H1) or (H2) be valid, the outcome for \textit{Jánošík} may be biased by an unwanted genre or authorial heterogeneity in Steigerwald’s training data. This effect is quite unlikely, as we have seen no evidence of it in classification of Steigerwald’s theatre plays during cross-validation; to rule out this possibility, however, we perform the classification once again, this time excluding \textit{Sun} from the training set. The results remain unchanged: all models still predict Kundera as author. 

On the other hand, should (H3) be valid, \textit{Jánošík}---as allegedly the earliest of Steigerwald’s plays---may be a case of a play written under the even stronger influence of Kundera, and therefore misclassified as such. This is even more unlikely, as there were just two years between the completely {Kunderian} \textit{Jánošík} and fairly recognisable \textit{Sun}. This strikes us as too short an interval for such a significant shift in authorial style. Yet, to take into account this possibility, we take a closer look at our datasets by means of hierarchical clustering.

\begin{figure*}[!t]
  \centering
      \includegraphics[width=0.85\textwidth]{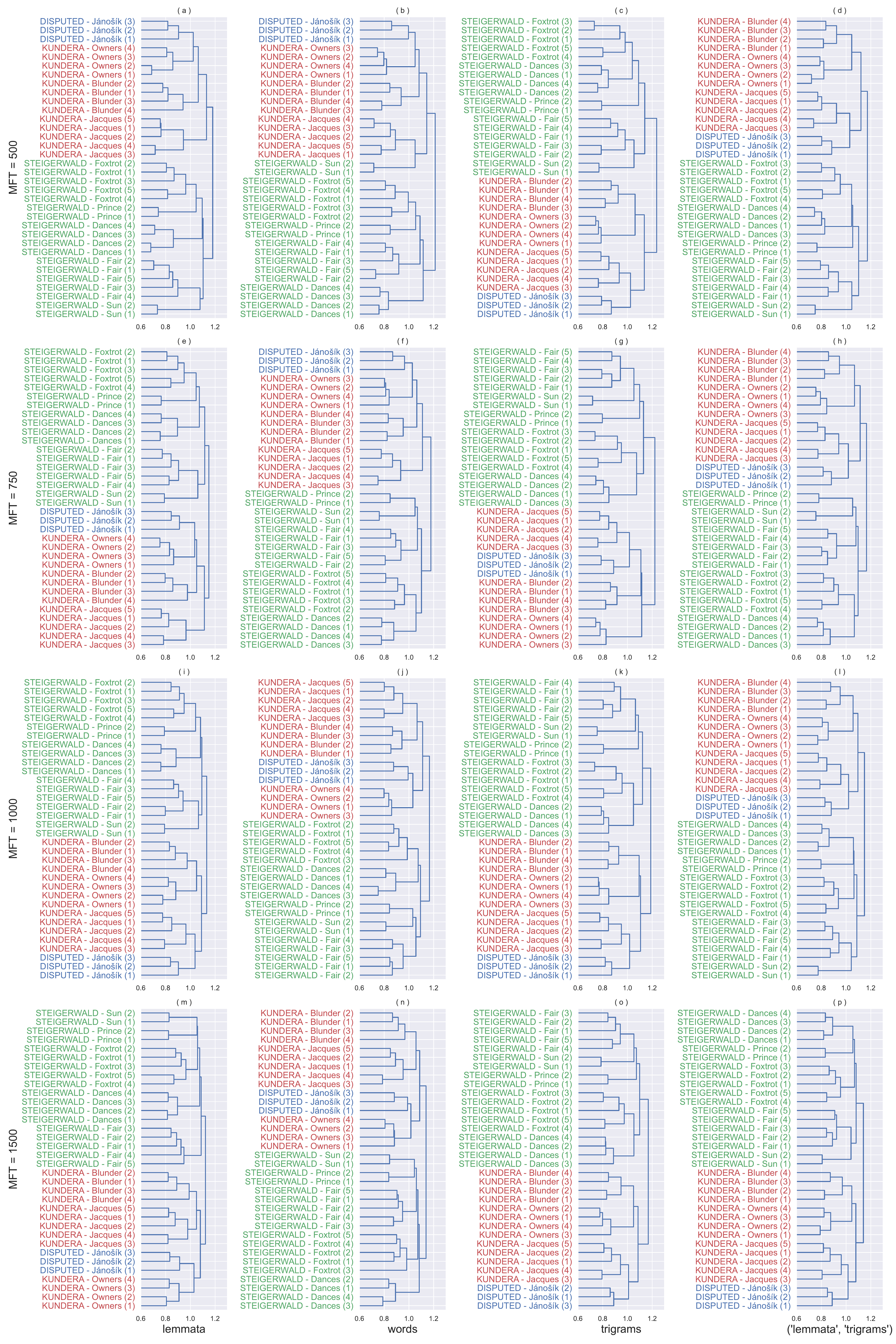}
  \caption{Dendrograms based on different feature sets and different levels of Most Frequent Types (cosine distance, complete linkage)}
  \label{fig:dendro}
\end{figure*}

If the above situation is the case, we may expect \textit{Jánošík} and \textit{Sun}---as Steigerwald’s strongly and mildly \say{Kunderian} plays, respectively---to tend to cluster together. \autoref{fig:dendro} however, gives a completely different picture. The two never appear alongside one another. In one case (words, 500 MFT), Sun clusters with Kundera’s Jacques, but there is a long way through the tree from these to \textit{Jánošík}. In all other cases the two uppermost clusters perfectly separate the data to Steigerwald and Kundera, with \textit{Jánošík} always appearing in the latter. The lowermost clustering of \textit{Sun} is quite intuitive: it mostly appears alongside \textit{Fair}---the play by Steigerwald closest to it on the timeline. For \textit{Jánošík}, it is a similar case with trigrams and lemmata+trigrams combinations (attracted to \textit{Jacques}, the closest play by Kundera on the timeline), yet with the remaining two features it tends to find its nearest neighbour in Kundera’s earliest work, \textit{Owners}. We thus find no evidence supporting (H3) and lean towards the simplest explanation for \textit{Jánošík} being constantly assigned to Kundera: that he is its actual author.

\section{Conclusions}

We believe enough evidence has been gathered here to claim that the real author of \textit{Juro Jánošík} is Milan Kundera. It should added, in conclusion, that Kundera’s biographer Jan Novák claims that this is not the only play secretly published by the author. After 1969, when he was let go from his job and prohibited from publishing his work, Kundera most likely found himself in an existential crisis, not knowing what to do next or how to make ends meet. He was therefore compelled to publish under a foreign name in television and radio – media for which he had previously expressed disdain – turning once more to a former student for help, this time the Slovak Natália Ivančová, who offered his plays for young audiences to Czechoslovak Television Bratislava. They were presented under the name of Peter Ševčovič, a dramaturge working on the editorial board for children’s programming at ČST Bratislava (Ševčovič also translated Kundera’s scripts from Czech to Slovak).\footnote{Natália Ivančová confirmed this fact to Kundera’s biographer; see \cite[781ff]{novak2020}.} \textit{Simultaneous with Aľjochin} (\textit{Simultánka s Aljokhin}) presents the story of a Slovak boy who, after a Nazi and enthusiastic chess player moves in with them, stops playing chess. The second script, intended for a three-part series called \textit{A friend for both of us} (\textit{Kamarádka pro nás dva}), depicts the process of finding a partner for a sympathetic father who democratically and sympathetically raises his teenage son.

Ivančová would go on to offer two more of Kundera’s plays to Czechoslovak Radio: \textit{Farewell, my love} (\textit{Sbohem, moje lásko}) is the story of a teenager who has a problematic relationship with his father. When he identifies a thief in the store where his father works, their relationship improves. \textit{Mahler’s symphony} (\textit{Mahlerova symfonie}) is a mysterious tragedy from contemporary Spain in which a judge’s wife glimpses the impending murder of her husband and herself. While the first of these radio plays was recorded for radio in 1975, the second was never realised, allegedly because the true identity of the author was revealed.

\section*{Acknowledgment}
Data and code are available at \url{https://github.com/versotym/kundera}. Since all works in our corpus are copyright protected, we publish frequency tables only, from which reconstruction of original texts is not possible.

We would like to thank to Milan Straka (Institute of Formal and Applied Linguistics, Charles University) for kindly performing an experiment for our needs with lemmatisation of punctuation stripped texts.


\printbibliography

\end{multicols}
\end{document}